# Graph based E-Government web service composition

Hajar Elmaghraoui [1], Imane Zaoui[2], Dalila Chiadmi[3] and Laila Benhlima[4]

[1] Department of Computer Science, Mohammad Vth University -Agdal, Mohammadia School of Engineers(EMI)
Rabat, Agdal, Morocco

[2] Department of Computer Science, Mohammad Vth University -Agdal, Mohammadia School of Engineers(EMI)
Rabat, Agdal, Morocco

[3] Department of Computer Science, Mohammad Vth University -Agdal, Mohammadia School of Engineers(EMI)
Rabat, Agdal, Morocco

[4] Department of Computer Science, Mohammad Vth University -Agdal, Mohammadia School of Engineers(EMI)
Rabat, Agdal, Morocco

## Abstract

Nowadays, e-government has emerged as a government policy to improve the quality and efficiency of public administrations. By exploiting the potential of new information and communication technologies, government agencies are providing a wide spectrum of online services. These services are composed of several web services that comply with well defined processes. One of the big challenges is the need to optimize the composition of the elementary web services. In this paper, we present a solution for optimizing the computation effort in web service composition. Our method is based on Graph Theory. We model the semantic relationship between the involved web services through a directed graph. Then, we compute all shortest paths using for the first time, an extended version of the Floyd-Warshall algorithm.

*Keywords: Web services composition, optimization, graph theory; Floyd-Warshall, e-government.*

## 1. Introduction

Many countries around the world are attempting to strengthen and revitalize the quality and the efficiency of their public administration and make it more service oriented. In the last decade, electronic (e-) government has emerged as a solution to the problems of traditional public administrations such as high costs, poor quality services and corruption. According to the European Commission, e-government is "the use of Information and Communication Technologies (ICTs), in public administrations combined with organizational change and new skills in order to improve public services and democratic processes and strengthen support to public policies" [1]. Thus, ICT integration in government operations plays a crucial role in improving the quality and transparency of public services in several domains including social programs, healthcares, tax filling, voting, etc… ICT has then become crucial to a successful e-Gov program. By deploying public services via the Internet, communication is made easier for citizens and businesses, resources are federated, and finally are sped up substantially. Furthermore, adopting web services in e-Gov enables government agencies to provide value-added services through the service composition process. Web service composition allows flexible creation of new services by assembling independent and reusable service components. Traditionally, web services are composed either in a static or dynamic ways each time a user asks for a service. These methods are still tedious and very costly. Thus, one of the big challenges in e-service composition is optimizing the composition effort. By composing the most suitable web services with the lowest costs, we will certainly increase e-government efficiency, reduce considerably the response time and give more satisfying responses to users' queries. In this paper, we propose a graph based approach for optimizing web service composition. The approach is based on representing the web services semantic relationship using a directed graph built at the time of publishing. This graph is traversed to find the optimized combination of all component services that compose targeted services. The solution extends the Floyd-Warshall algorithm to reconstruct all shortest paths between all the vertices in the service graph. This operation is performed at the time of web service publishing. Indeed, by computing shortest paths before executing user queries, we optimize the composition time and costs.

The rest of this paper is organized as follows. In section 2, we present some related work. Section 3 outlines our optimization approach .In section 4, we present our web services model which is based on the graph theory. Section 5 describes our solution for optimizing web service composition, based on Floyd-Warshall algorithm. We illustrate our approach in section 6 with an example of retiring e-services. We conclude in section 7 with ongoing works.

## 2. Related Work

Many proposals of web services composition methods have been presented in recent years. For a detailed survey, we refer to [2, 3]. In this section, we present a brief overview of some techniques that deal with automatic web service composition. We consider only techniques that use service dependency information, graph models, and semantics. The simple idea behind dependency is that whenever a web service receives some input and returns some output, the output is somehow related or dependent on the given input. By using a graph model, the behavior of available web services is represented in terms of their input-output information, as well as semantic information about the web data. A graph is a collection of vertices or 'nodes' and a collection of edges that connect pairs of vertices. A graph may be undirected, meaning that there is no distinction between the two vertices associated with each edge, or its edges may be directed from one vertex to another. A weighted graph is a graph where each edge has a weight (some real number) associated with it. The dependency graph is used in finding a composite service to satisfy a given request.

Most of composition graph-based methods build web services dependency graphs at runtime. They use a search algorithm for traversing dependency graphs in order to compose services. The main difference between these methods is attributed to how they search the dependency graph. A*, Dijkstra, Floyd, Forward chaining, backward chaining and bidirectional search algorithms are examples of the most common search techniques.

Hashemian et al. [4] store I/O dependencies between available Web services in their dependency graph, and then build composite services by applying a graph search algorithm. In their graph, each service and I/O parameter is represented as a vertex, service's input and output are represented as incoming and outgoing edges, respectively. The authors consider only the matching and dependencies between input and output parameters without considering functional semantics, thus they cannot guarantee that the generated composite services provides the requested functionality correctly.

In [5], the authors use the backward chaining method in combination with depth first search to get the required services for a composite task. Their solution is rather abstract and does not clearly discuss execution plan generation algorithm.

Arpinar et al. [6] present an approach which not only use graphs for web service composition, but also use semantic similarity as we present in this work. They consider edges with weights and deploy a shortest-path dynamic programming algorithm based on Bellman-Ford's algorithm for computing the shortest path. For cost, the authors consider the execution time of each service and input/output similarity but they don't take into consideration the services' nonfunctional attributes.

Gekas et al. [7] develop a service composition registry as a hyperlinked graph network with no size restrictions, and dynamically analyze its structure to derive useful heuristics to guide the composition process. Services are represented in a graph network and this graph is created and explored during the composition process to find a possible path from the initial state to a final state. In order to reduce the time of searching, a set of heuristics is used. But according to the authors, creating the graph at the time of composition is very costly in term of computation and limits the applicability of graph-based approaches to the problem of web service composition.

Talantikite et al. [8] propose to pre-compute and store a network of services that are linked by their I/O parameters. The link is built by using semantic similarity functions based on ontology. They represent the service network using a graph structure. Their approach utilizes backward chaining and depth-first search algorithms to find sub-graphs that contain services to accomplish the requested task. They propose a way to select an optimal plan in case of finding more than one plan. However, they also create the graph at the time of composition which incurs substantial overhead.

In this paper, we propose a graph based solution which creates the graph at the time of publishing and therefore optimize the composition by reducing the computational effort at the time of composition. Our approach uses the Floyd-Warshall algorithm.

# 3. The optimization approach

Conceptually, our problem can be described as follows: "Given a set of available services, and given a goal, our aim is to automatically compose an optimal subset of services to satisfy the goal". We propose a solution based on Graph Theory. Our optimization approach consists of two fundamental pillars: i) representing the semantic relationships between all available web services using an oriented graph called SCG (Service Composition Graph), and ii) Applying a graph search algorithm in order to compute all shortest paths between all nodes. We store the algorithm results into a matrix called the Shortest Predecessor Matrix (SPM). SPM is used to reconstruct the shortest path which corresponds to an optimal automated service composition. In order to find the best combination of web services that meets the user goal, we propose to extend the Floyd Warshall graph search algorithm.

The SCG and the SPM are built at the time of publishing and updated each time new services are published or existing services are changed or removed from the repository. Thus, we avoid building the graph and applying the graph search algorithm each time a request for a service composition is made. By creating the graph and computing the shortest paths at the time of publishing, we reduce considerably the computational effort at the time of composition and thereafter the execution time of composition.

During the composition runtime, and for each query, we identify the starting and ending web services (called Vstrat and Vgoals respectively) into the SCG. Then, we extract from the SPM matrix the corresponding shortest path between VStart and Vgoals. Our optimization approach is summarized in the following process represented in Figure 1.

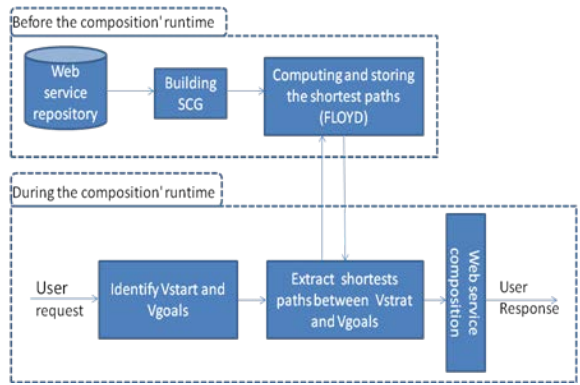

Fig.1. Optimization process

The details about this process are provided in the next sections.

# 4. Modeling web service composition as a directed weighted graph

In this section, we present the service composition graph (SCG).

## 4.1 Defining the service composition graph (SCG)

We start from a local Semantic Web Services repository, which is populated with OWL-S [9] descriptions (a popular and well-understood solution to support ontology-based Semantic Web). We will use the semantic description of the input and output parameters of the services to build a Service Composition Graph (SCG). Thus, building a SCG is based on semantic similarity between web services. In fact, for a set of web services, we need to check if two services are to be invoked in sequence during the composition process. This means that for each input of the former service, there is some output of the following service that is equivalent, more, or less general than the demanded input. In such cases, we can say that these services are semantically composable.

Formally, the SCG is a directed Graph G= (V,E,W), where V is the set of vertices representing the web services, E is the set of edges representing the semantic relationship between web services, and W is the set of edges's weights. To create the SCG, we follow the following procedure:

- For each web service $WS_i$ in the repository, create a vertex $V_i$ in the graph
- If two web services $WS_i$ and $WS_j$ represented respectively by vertices $V_i$ and $V_j \in V$ have a semantic similarity among their inputs and outputs, then introduce an edge $V_i \rightarrow V_j$
- For each edge connecting vertices $V_i$ and $V_j$, associate a weight $W_{ij}$

## 4.2 Measuring the semantic similarity

Semantic similarity stands for the degree of likeness between concepts. To compute the semantic similarities among services, we use subsumption reasoning as originally proposed by Paolucci et al. [10]. Subsumption reasoning verifies whether a concept is more general than another one. Given two web services represented by vertices $V_i$ and $V_j$, this reasoning allows computing the degrees of similarity between the services using the scale: equivalent, subclass, subsumes and the following rules:

**1-Exact match**: If the output parameters of $V_i$ and the input parameters of $V_j$ are equivalent concepts.
**2-Plug-in match**: If the output of $V_i$ is a sub-concept of the input of $V_j$ ($V_j$ subsumes $V_i$)
**3-Subsumes match**: If the input of $V_j$ is a sub-concept of the output of $V_i$. ($V_i$ subsumes $V_j$)
**4-Fail match**: No subsumption nor equivalence relation between $V_i$ and $V_j$.
Thus, we associate an edge connecting vertices $V_i$ to $V_j$ if the degree of similarity between the outputs of $V_i$ and the inputs of $V_j$ is Exact, PlugIn or Subsumes.

### 4.3 Weighting the SCG

The weight is a key point of the model. It influences the choice of composition paths which directly affects the composition result. We associate to all the edges in the SCG a weight calculated using the semantic similarity value S between input and output parameters, and a function f(QOS) of non-functional properties of services that are cost, execution time, reliability, availability…etc.
In this paper, we propose three criteria as parameters for f(QOS) which are: cost, time and availability where:

- **Cost:** the fee that a requester has to pay for invoking the service $V_i$.
- **Execution time:** measures the expected delay time between the moment when $V_i$ is invoked and when the results are received.
- **Availability:** is the probability that $V_i$ is accessible.

Other non-functional properties can be considered in computing the weight of the edges, such as reliability, reputation, security…etc.

We calculate f(QOS) as follows:

$f(QOS(V_i))=(\alpha*cost)+(\beta*execution\ time)+(\mu*availability)$  (1)

Where $\alpha$, $\beta$ and $\mu$ are relative factors that can be defined by the system administrator.
The weight $W_{ij}$ of a given edge $V_i \rightarrow V_j$ is computed as follows:

$$W_{ij} = f(QOS(V_i)) + S_{ij} \qquad (2)$$

Where $S_{ij}$ is the semantic similarity value between the input parameters of $V_j$ and the output parameters of $V_j$.

## 5. Optimizing the web service composition

Given a user query which specifies web service's inputs and outputs, the composition problem involves automatically finding a directed acyclic graph of services from the SCG that can be composed to get the required service, when a matching service is not found. Our service composition research aims at reducing the complexity and time needed to generate and execute a composition. We also improve its efficiency by selecting the best possible services available. To achieve these optimization goals, we compute at the time of publishing all shortest paths between every pair of vertices in the SCG using the Floyd all-pairs shortest path which is a dynamic programming algorithm. Then, for each user query, we identify in the SCG the start and goal vertices based on semantic similarities between input and output parameters of the query and the SCG vertices. Thus, we calculate the shortest path between the start and goals vertices which represent the optimal services combination that meet the user needs.

### 5.1 Identifying start and goal vertices

Given the weighted directed graph SCG that models the web services in our repository, and when we receive a user query that requires service composition, we identify vertices and edges which represent the input and output parameters provided by the requester and we update temporarily the graph as follows:

- A starting node ($V_{START}$) is created and connected with all vertices (services) that contain at least one input provided by the requester;
- For each output demanded by the requester, create a goal node ($V_{GOAL}$) and connect it with all services providing this output.

These additional nodes are used to guide the service composition, which is based on the computation of minimum cost paths which are the shortest paths from the start node (representing the inputs) to the goal nodes (representing the outputs).

### 5.2 Shortest path issue

The shortest path problem has been widely used in many fields such as project planning, geographic information systems and military operations research. The classical algorithms to solve the shortest path problem are mainly Floyd algorithm and Dijkstra algorithm. Floyd algorithm is a multi-source shortest path algorithm, which is mainly used to calculate the shortest path among all nodes; whereas Dijkstra algorithm is a single-source shortest path algorithm, which is an efficient algorithm, used to calculate the shortest path from a source node to its all places nodes.

In order to reduce the overhead at runtime, we need to calculate the shortest paths for all the vertex pairs in advance and store this information for an eventual use at the time of service composition. Of course, this problem can be solved by applying Dijkstra's algorithm, but the program will be complex and the computational overhead will be large to O(N⁴) [11], where N is the number of vertices. However, if we use Floyd algorithm, the computation will be simplified with a time complexity of O(N³) [11]. Floyd's algorithm is globally optimal, and the code is simple, easy to implement, and easy to integrate with other program modules [12]. In the next section, we will present a brief description of this algorithm.

Once the user request is received, we define the start node ($V_{START}$) and the set of goal nodes ($V_{GOAL}$) introduced in section 4.1, then we extract from the pre-computed list of shortest paths (already calculated using the Floyd algorithm and stored for future use), the shortest paths from the node ($V_{START}$) to each goal node member of ($V_{GOAL}$). This means that paths are found from the available input parameters, to the desired output parameters. Thus, we generate partial compositions that may have common vertices. Since our goal is to generate a single connected graph, we achieve this by eliminating duplicate paths by analyzing the intersections of the extracted paths.

## 5.3 Overview of Floyd- Warshall algorithm

The Floyd-Warshall algorithm [13] is an efficient Dynamic Programming [14] algorithm that computes the shortest paths between every pair of vertices in a weighted and potentially directed graph. This is arguably the easiest-to-implement algorithm for computing shortest paths in the literature [15]. The time complexity of this algorithm takes O(N³). A single execution of the algorithm will provide the lengths (summed weights) of the shortest paths between all pairs of vertices though it does not return details of the paths themselves. With some modifications of the algorithm, we create a method to reconstruct the actual shortest path between any two endpoint vertices. Path reconstruction runs in O(**N**) and thus the complexity of Floyd-Warshall algorithm is not affected. We describe below Floyd-Warshall algorithm with path reconstruction.

- Let DIST be a NxN adjacency matrix (N is the number of vertices), DIST(i,j) representing the length (or cost) of the shortest path from $V_i$ to $V_j$. For each element DIST(i,j) assign a value equal to the cost of the edge going from Vi to Vj, or an infinity value if this edge does not exist.
- At each step, for each pair of vertices $V_i$ and $V_{j,}$ check if there is an intermediate vertex $V_k$ so that the path from $V_i$ to $V_j$ through $V_k$ is shorter than the one already found for $V_i$ and $V_j$.
- Let NEXT be an N×N matrix that will contain the final path. The NEXT matrix is updated along with the DIST matrix such that at completion both tables are complete and accurate, and any entries which are infinite in the DIST table will be null in the NEXT table. The path from $V_i$ to $V_j$ is then the path from $V_i$ to NEXT(i,j), followed by path from NEXT(i, j) to $V_j$.

In the next section, we present briefly the pseudo code of our Floyd-Warshall algorithm with path reconstruction.

## 5.4 Pseudo code of Floyd- Warshall with Path Reconstruction

# We assume an input graph of N vertices
# weight (i,j) is the weight of the edge from vertex $V_i$ to $V_j$
: equal to infinity if such an edge does not exist and 0 if i=j

```
BEGIN
For i = 1 to N
   For j = 1 to N
      #Initialization of the adjacency matrix
      DIST (i,j) = weight(i,j)
      # Initialization of the predecessor matrix
      If i! =j and there exists an edge from i to j
      Then
            NEXT (i,j) = i
      Else
            NEXT (i,j) = NIL
For k=1 to N                    # k is the intermediate vertex
   For i=1 to N
      For j=1 to N
# check if the path from i to j that goes through k is shorter
than the one already found
      If     DIST(i,k) + DIST(k,j) < DIST(i,j)
      Then
            # New shorter path length
            DIST (i,j) = DIST(i,k) + DIST(k,j)
            # The predecessor matrix
            NEXT (i,j)=k

Return DIST   # matrix of final distances
```

// GetPath(i,j) : function for path reconstruction between two vertices $V_i$ and $V_j$.

```
If        DIST (i,j) = infinity
Then
          Return "no path"
IF        NEXT (i,j)= Null
Then
          Return " "   # there is an edge from $V_i$ to $V_j$, with
no vertices between
Else
          Return        GetPath(i,NEXT(i,j))+NEXT(i,j)+
GetPath(NEXT(i,j),j)
END.
```

While the outcomes of our research are generic enough to be applicable to a wide range of applications, we use the area of e-government as a case study.

# 6. Case study

In this section, we illustrate our approach for web services composition with the example of retiring web services offered by the RCAR agency (Regime Collectif des Allocations de Retraite). The agency allows all members to access the retirement services via its web portal. The portal www.rcar.ma contains four spaces which are employer space, member space, recipient space and provider space. This categorization customizes the agency services according to different user profiles. All web services are stored in the repository. We model the relationship between them using a SCG as we have explained in section 3.

## 6.1 The retiring SCG

The SCG of the RCAR contains a hundred of web services. To simplify, we only present in Figure.2, a part of the SCG of the RCAR agency. Vertices are the web services described in table 1, edges represent the relationships between inputs and outputs and the weights represent the web services costs. Note that for confidentiality issues, web services and values are purely illustrative.

Table 1: Web services description

|       | Name       | Description                                                                 |
|-------|------------|-----------------------------------------------------------------------------|
| $V_1$ | GetAuth    | Allows authenticating users using a login and password                      |
| $V_2$ | FillForm   | Return and fill adequate form for the demanded service                      |
| $V_3$ | GetRegister| Allows new users to subscribe in order to get their login and passwords     |
| $V_4$ | SendReq    | Send the user's request with the input data to the RCAR system              |
| $V_5$ | FillStatus | Allows users to create their status during the registration process and to update it. |
| $V_6$ | SendResp   | The system send the adequate response to the user's request                 |
| $V_7$ | Unsubscribe| Allows user to unsubscribe, which automatically delete his profile and status |
| $V_8$ | CheckData  | Checks the user information to get                                           |

|  |  |
|--|--|
|  | the adequate response. |

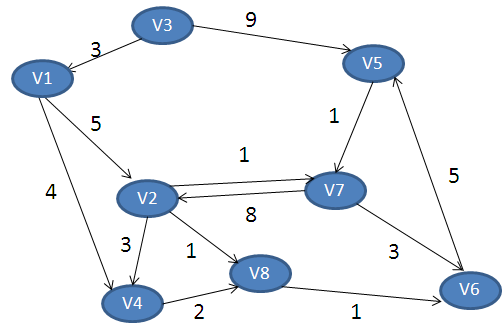

Fig.2. The retiring SCG

## 6.2 Shortest paths using Floyd-Warshall with path reconstruction

We apply the Floyd-Warshall algorithm with path reconstruction presented in section 4.4 to compute all shortest paths between all nodes in the SCG. The adjacency matrix, which represents the weights between all pairs of vertices in the SCG, is given bellow (3).

$$
\begin{array}{c|cccccccc}
 & V_1 & V_2 & V_3 & V_4 & V_5 & V_6 & V_7 & V_8 \\
\hline
V_1 & 0 & 5 & \infty & 4 & \infty & \infty & \infty & \infty \\
V_2 & \infty & 0 & \infty & 3 & \infty & \infty & 1 & 1 \\
V_3 & 3 & \infty & 0 & \infty & 9 & \infty & \infty & \infty \\
V_4 & \infty & \infty & \infty & 0 & \infty & \infty & \infty & 2 \\
V_5 & \infty & \infty & \infty & \infty & 0 & \infty & 1 & \infty \\
V_6 & \infty & \infty & \infty & \infty & 5 & 0 & \infty & \infty \\
V_7 & \infty & 8 & \infty & \infty & \infty & 3 & 0 & \infty \\
V_8 & \infty & \infty & \infty & \infty & \infty & 1 & \infty & 0 \\
\end{array} \quad (3)
$$

The Shortest Predecessor Matrix (SPM) that will be used to extract all shortest paths, given bellow (4), will be stored to be manipulated at the time of service composition. Note that $(V_i, V_j)$ holds a vertex $V_k$ which is the direct predecessor of $V_j$ on the least cost path between $V_i$ and $V_j$

$$
\begin{array}{c|cccccccc}
 & V_1 & V_2 & V_3 & V_4 & V_5 & V_6 & V_7 & V_8 \\
\hline
V_1 & \varnothing & V_1 & \varnothing & V_1 & V_3 & V_8 & V_2 & V_2 \\
V_2 & \varnothing & \varnothing & \varnothing & V_2 & V_7 & \varnothing & V_2 & V_2 \\
V_3 & V_3 & V_1 & \varnothing & V_1 & V_3 & V_7 & V_2 & V_2 \\
V_4 & \varnothing & \varnothing & \varnothing & \varnothing & V_6 & V_8 & \varnothing & V_4 \\
V_5 & \varnothing & V_7 & \varnothing & V_7 & \varnothing & V_7 & V_5 & V_7 \\
V_6 & \varnothing & V_7 & \varnothing & V_7 & V_6 & \varnothing & V_6 & V_7 \\
V_7 & \varnothing & V_7 & \varnothing & V_2 & V_6 & V_7 & \varnothing & V_2 \\
V_8 & \varnothing & \varnothing & \varnothing & \varnothing & V_6 & V_8 & \varnothing & \varnothing \\
\end{array} \quad (4)
$$

We note also that in case of adding, deleting or modifying the services, the SCG will be updated as well as the adjacency matrix (3) and the Shortest Predecessor Matrix (SPM) (4). This update guarantees an accurate web service composition.

### 6.3 Delivering the retiring e-services based on the optimized web service composition.

Let us consider a member seeking to get an online certificate of his membership ($V_{GOAL1}$), to calculate his retirement pension ($V_{GOAL2}$) and to update his profile information ($V_{GOAL3}$). All these operations require user registration and authentication. These inputs are related to the start node ($V_{START}$). To get his request satisfied, several web services which are transparent for the user are involved. These web services are $V_1$: *GetAuth*, $V_4$: *SendReq*, $V_8$:*CheckData, and $V_6$:SendResp*. As explained before, the first step of our web service composition is to identify the root ($V_{START}$) and the targeted nodes $V_{GOAL1}$, $V_{GOAL2}$ and $V_{GOAL3}$ that answer the user request. The identification is based on comparing the inputs and outputs of the existing web services with the user's query. The result of this operation is connecting virtual vertices with a null weight to the SCG. Figure 3 illustrates this stage.

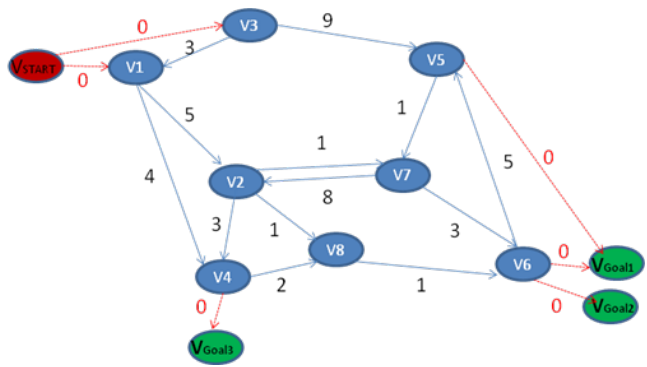

Fig.3. Identifying the START AND GOALS Vertices into the SCG

After identifying the input and output services in the SCG, which are connected to $V_{START}$ and $V_{GOAL1}$, $V_{GOAL2}$ and $V_{GOAL3}$, the second step of the web service composition is to extract from the matrix (2), all shortest paths between $V_{START}$ and all $V_{GOALS}$. The results are given in table 2 and the optimal sub-graph of services that will meet the user needs is illustrated by figure 4.

Table 2: Shortest paths for web service composition

| | |
|---|---|
| $V_{START}$ to $V_{GOAL1}$ | V1→V4→V8→V6→V$_{GOAL1}$ |
| $V_{START}$ to $V_{GOAL2}$ | V1→V4→V8→V6→V$_{GOAL2}$ |
| $V_{START}$ to $V_{GOAL3}$ | V1→V4→V$_{GOAL3}$ |

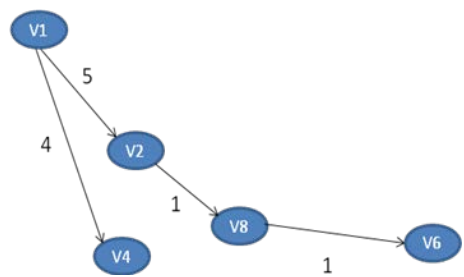

Fig.4. the optimal sub-graph of services that respond to the user request

## 7. Conclusion and ongoing work

Automatic composition of web services has drawn a great deal of attention recently. By composition, we mean taking advantage of currently existing web services to provide a new service that does not exist on its own. In this paper, we present a graph based approach for optimizing web service composition, which takes into consideration the semantic similarity of services computed using subsumption reasoning on their inputs and outputs. Our results include:

- Modeling the web service composition by means of a directed weighted graph, where the weight calculation takes into account the non-functional properties of services and the semantic similarity between them.

- Using Floyd algorithm to compute the shortest paths at the time of publication, in order to reduce the complexity and time needed to generate and execute a web service composition.

The implementation and evaluation of the solution proposed in this paper in real systems is the main focus of our ongoing work. We also intend to evaluate other cost policies such as reliability, reputation, security…etc.

Our near future work is mainly focusing on addressing reliability and availability of web services. Indeed, during the execution of web service composition, if one service fails or becomes unavailable, a failure recovery mechanism is needed to ensure that the running process is not interrupted and the failed service can be replaced quickly and efficiently.